\documentclass[conference]{IEEEtran}
\IEEEoverridecommandlockouts
\usepackage{subcaption}
\usepackage{cite}
\usepackage{amsmath,amssymb,amsfonts}
\usepackage{algorithmic}
\usepackage{graphicx}
\usepackage{multirow}
\usepackage{xcolor}
\usepackage{enumitem}
\usepackage{textcomp}
\usepackage{xcolor}
\def\BibTeX{{\rm B\kern-.05em{\sc i\kern-.025em b}\kern-.08em
    T\kern-.1667em\lower.7ex\hbox{E}\kern-.125emX}}
\begin{document}

\title{Privacy-Preserving Clothing Classification using Vision Transformer for Thermal Comfort Estimation\\
}

\author{
    \IEEEauthorblockN{Tatsuya Chuman}
    \IEEEauthorblockA{\textit{NTT FACILITIES, INC.} \\
    Tokyo, Japan \\
    tatsuya.chuman.mg@ntt-f.co.jp}
    \and
    \IEEEauthorblockN{Yosuke Udagawa}
    \IEEEauthorblockA{\textit{NTT FACILITIES, INC.} \\
    Tokyo, Japan \\
    udagaw25@ntt-f.co.jp}
    \and
    \IEEEauthorblockN{Hitoshi Kiya}
    \IEEEauthorblockA{\textit{Tokyo Metropolitan University} \\
    Tokyo, Japan \\
    kiya@tmu.ac.jp}
}

\maketitle

\begin{abstract}
A privacy-preserving clothing classification scheme is presented to enable secure occupant-centric control (OCC) systems.
Although the utilization of camera images for HVAC control has been widely studied to optimize thermal comfort, privacy protection of occupant images has not been considered in prior works. While various privacy-preserving methods have been proposed for image classification, applying conventional schemes results in severe accuracy degradation.
In this paper, we introduce a privacy-preserving classification method using Vision Transformer (ViT) applied to clothing insulation estimation.
In an experiment using the DeepFashion dataset categorized by clothing insulation, while the conventional pixel-based method suffers a severe accuracy drop, our scheme maintains a high accuracy on encrypted images, showing no degradation from plain images across all categories.
\end{abstract}


\section{Introduction}
The use of camera images in building management systems has greatly increased because of the rapid growth of deep learning and the demand for energy-efficient heating, ventilation, and air conditioning (HVAC) control. In particular, occupant-centric control (OCC), which utilizes occupant information such as locations and activities, has been widely studied to optimize thermal comfort.
This optimization often relies on the Predicted Mean Vote (PMV) index, in which clothing insulation is recognized as a critical parameter that substantially impacts occupant thermal comfort~\cite{fanger1970thermal}.

While many studies on image-based HVAC control have been reported, most of them focused on control logic, without considering the privacy protection of occupant images~\cite{choi2022deep}. Moreover, with image classification often performed on the cloud, transmitting raw images that include personally identifiable information (PII) to an untrusted provider raises significant privacy concerns. Therefore, a privacy-preserving image-based control scheme is essential.
However, there is a trade-off between privacy and the performance of signal processing in the encrypted domain. In conventional privacy-preserving methods utilizing Convolutional Neural Networks (CNNs), the pixel-based encryption leads to significant accuracy degradation \cite{sirichotedumrong2019privacy}.

In this paper, we present a novel clothing classification scheme utilizing a Vision Transformer (ViT) model, such as ViT-S/16, with a key-based image encryption and model transformation \cite{kiya2022image}.
Experimental results on four clothing categories from the DeepFashion dataset~\cite{liu2016deepfashion} demonstrated that the proposed scheme achieves high classification accuracy on encrypted images, which is significantly superior to the conventional method, making it highly effective for privacy-preserving HVAC control systems and PMV calculation.

\begin{figure}[t]
  \centering
  \includegraphics[width=0.75
\linewidth]{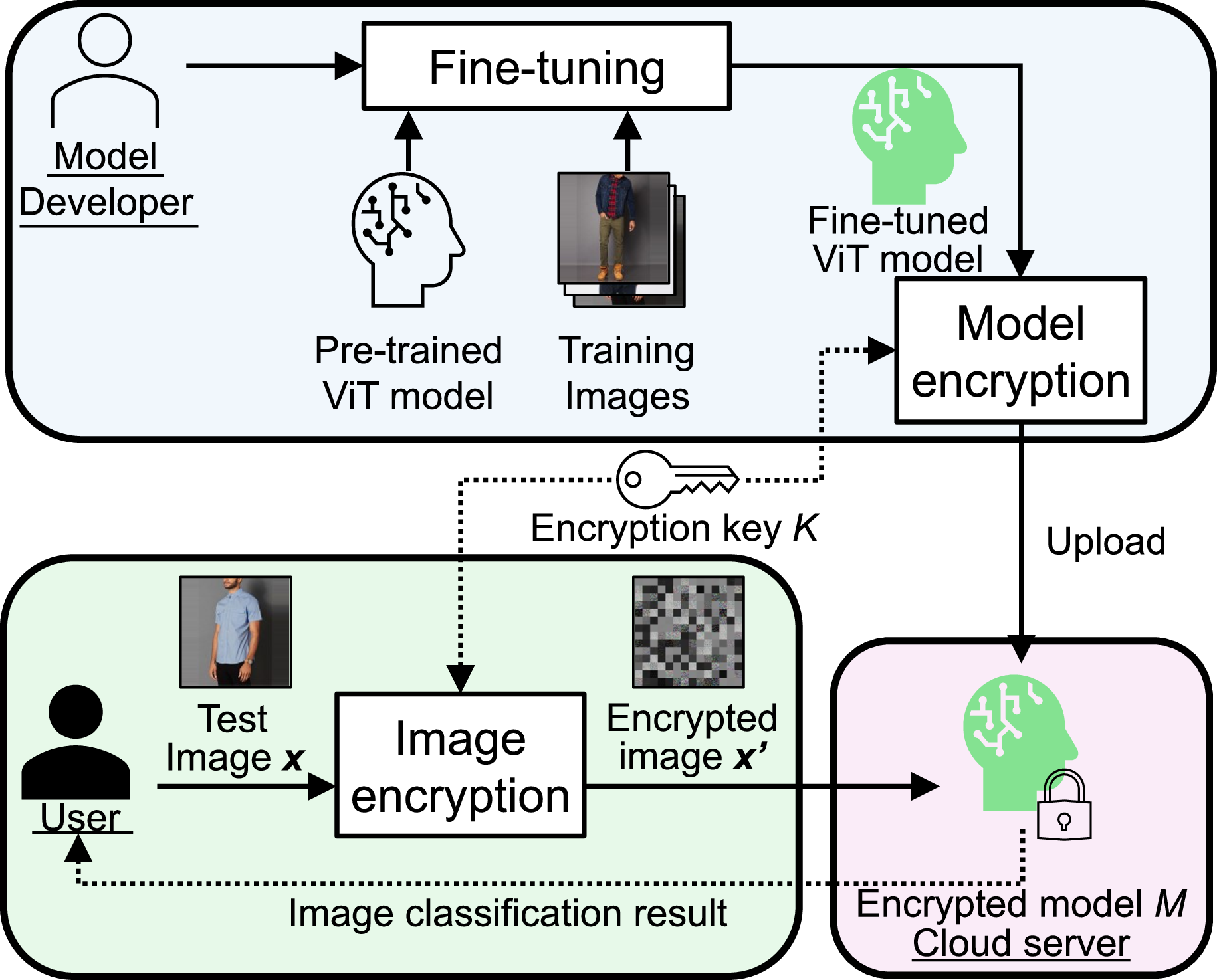}
  \caption{Overview of the privacy-preserving scheme.}
  \label{fig:system_overview}
  \vspace{-0.4cm}
\end{figure}

\section{Proposed Method}
\subsection{Overview of the Proposed Scheme}
Figure \ref{fig:system_overview} illustrates our clothing estimation scenario. Image classification involves a model developer and a user. First, the developer constructs the model through the following steps:
\begin{enumerate}[label=(\arabic*), nosep, leftmargin=*]
    \item Fine-tuning a pre-trained ViT model, such as ViT-S/16, using a clothing dataset.
    \item Encrypting the model parameters with a secret key $K$ to create an encrypted model $M$.
    \item Uploading $M$ to a cloud server.
\end{enumerate}
Next, the user obtains the classification result through the following steps:
\begin{enumerate}[label=(\arabic*), nosep, leftmargin=*]
    \item Encrypting a test image $\mathbf{x}$ with the shared key $K$ to generate $\mathbf{x}'$.
    \item Transmitting $\mathbf{x}'$ to the cloud server.
    \item Obtaining the classification result from the cloud server, which performs inference using the model $M$.
\end{enumerate}
This approach enables accurate classification while preserving the privacy of the input images, thereby exposing only visually protected data to the cloud server.

\subsection{Image Encryption for ViT}
\begin{figure}[t]
  \centering
  \begin{minipage}[b]{0.32\linewidth}
    \centering
    \includegraphics[width=\linewidth]{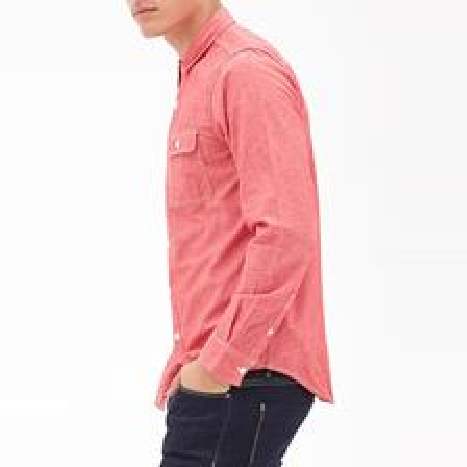}
    \subcaption{Original}
  \end{minipage}
  \hfill
  \begin{minipage}[b]{0.32\linewidth}
    \centering
    \includegraphics[width=\linewidth]{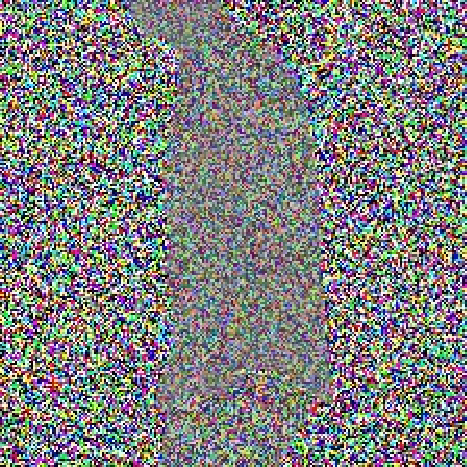}
    \subcaption{Pixel-based~\cite{sirichotedumrong2019privacy}}
  \end{minipage}
  \hfill
  \begin{minipage}[b]{0.32\linewidth}
    \centering
    \includegraphics[width=\linewidth]{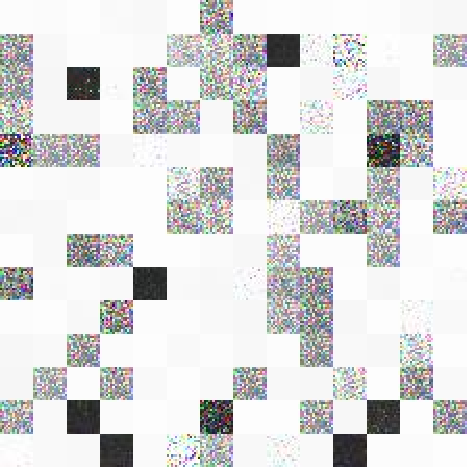}
    \subcaption{ViT-based~\cite{kiya2022image}}
  \end{minipage}
  
  \caption{Examples of encrypted images.}
  \label{fig:comparison}
  \vspace{-0.57cm}
\end{figure}

We employ an image and model transformation scheme with a secret key~\cite{kiya2022image}, which we refer to as the ViT-based scheme in this paper.
The input image $\mathbf{x}$ is divided into non-overlapping blocks of size $M \times M$, which corresponds to the patch size $P$ of the ViT model. Subsequently, a two-step encryption process is applied using pseudo-random integers generated by the key $K$.
First, pixel shuffling is performed, where pixels within each block are permuted using a common sequence.
Second, block scrambling is applied, where the blocks are spatially permuted to generate the encrypted image $\mathbf{x}'$.
Figures~\ref{fig:comparison}(b) and (c) are examples of images encrypted by the Pixel-based~\cite{sirichotedumrong2019privacy} and ViT-based~\cite{kiya2022image} schemes, respectively; Fig.~\ref{fig:comparison}(a) shows the original one.

Since ViT processes the image as a sequence of patches and its self-attention mechanism operates independently of the spatial arrangement, the model can extract features directly from $\mathbf{x}'$ without decryption, resulting in no accuracy degradation.

\vspace{-0.1cm}
\section{Experiments}
\subsection{Experimental Setup}
We evaluated the ViT-based scheme using the DeepFashion dataset~\cite{liu2016deepfashion}, which contains a total of 26,887 images.
The clothing labels were reorganized into four categories based on clothing insulation (clo-value), namely sleeveless, short-sleeve shirt, long-sleeve shirt, and outerwear including sweaters and jackets.
The number of images in each category is 11,033, 8,176, 4,218, and 3,586, respectively.
The dataset was randomly split into a training set (80\%) and a testing set (20\%).

Our experiments were conducted using the ViT-S/16 model, which was pre-trained on ImageNet-21k~\cite{deng2009imagenet}.
Input images were resized to $224 \times 224$ pixels to suit the input requirements of the ViT architecture.
Since the ViT-S/16 model uses a patch size of $P=16$, the encryption block size was set to $M=16$ accordingly.
We applied random horizontal flipping as the data augmentation technique during training.
For the optimization process, we employed the AdamW optimizer with a learning rate of $10^{-4}$, a weight decay of 0.01, and a batch size of 64, training the model for 50 epochs.

To evaluate the classification accuracy of the ViT-based scheme, we also utilized the pixel-based scheme~\cite{sirichotedumrong2019privacy} as a conventional method for comparison.

\subsection{Experimental Results}
Table~\ref{tab:accuracy_and_distribution} shows the classification accuracy of each model for plain and encrypted images.
The conventional pixel-based encryption applied to ResNet-18 caused a significant drop in overall accuracy, decreasing from 93.51\% to 83.34\%. This degradation was particularly severe in the long-sleeve shirt (67.40\%) and outerwear (65.64\%) categories. 
In contrast, the ViT-based scheme achieved an average accuracy of 95.65\% on encrypted images, showing no degradation compared to using plain images.
Furthermore, the ViT-based scheme outperformed the conventional pixel-based method across all clothing categories.

It was confirmed that the proposed scheme can be effectively applied not only to standard datasets but also to real-world applications, such as privacy-preserving clothing insulation estimation for PMV calculation.
Misclassification between clothing categories with different clo values may affect PMV estimation. Therefore, the proposed scheme, which maintains a 95.65\% accuracy on encrypted images, is effective for reliable PMV-based thermal comfort assessment.

\vspace{-0.1cm}
\section{Conclusion}
In this paper, we proposed a privacy-preserving clothing classification scheme utilizing ViT for occupant-centric HVAC control.
Experimental results indicated that the ViT-based scheme maintains a 95.65\% accuracy with no degradation from plain images, whereas the conventional pixel-based method shows a severe accuracy drop.
This study demonstrated the practical viability of the proposed scheme for real-world applications, enabling privacy-preserving clothing insulation estimation essential for PMV calculation.
The ViT-S/16 model, with 22.0M parameters, is also suitable for edge implementation in consumer electronics, further supporting the practical applicability of the proposed scheme.

\begin{table}[tb]
\caption{Comparison of classification accuracy (\%) between plain and encrypted images}
\label{tab:accuracy_and_distribution}
\centering
\footnotesize
\setlength{\tabcolsep}{6.4pt} 
\begin{tabular}{l|cc|cc}
\hline
\multirow{2}{*}{Category} & \multicolumn{2}{c|}{ResNet-18} & \multicolumn{2}{c}{ViT-S/16} \\ \cline{2-5}
 & Plain & Pixel-based~\cite{sirichotedumrong2019privacy} & Plain & ViT-based~\cite{kiya2022image} \\
\hline
Sleeveless           & 98.80 & 92.03 & 98.89 & \bf{98.89} \\
Short-Sleeve Shirt   & 96.24 & 87.44 & 96.31 & \bf{96.31} \\
Long-Sleeve Shirt    & 90.95 & 67.40 & 93.39 & \bf{93.39} \\
Outerwear            & 73.41 & 65.64 & 86.56 & \bf{86.56} \\
\hline
Overall              & 93.51 & 83.34 & 95.65 & \bf{95.65} \\
\hline
\end{tabular}
\vspace{-0.35cm}
\end{table}

\bibliographystyle{IEEEtran} 
\bibliography{IEEEabrv,references} 

@book{fanger1970thermal,
  author    = {Fanger, P. Ole},
  title     = {Thermal Comfort: Analysis and Applications in Environmental Engineering},
  publisher = {Danish Technical Press},
  address   = {Copenhagen, Denmark},
  year      = {1970}
}

@article{choi2022deep,
  author  = {Choi, Haneul and Jeong, Bonghoon and Lee, Joosang and Na, Hooseung and Kang, Kyungmo and Kim, Taeyeon},
  title   = {Deep-vision-based metabolic rate and clothing insulation estimation for occupant-centric control},
  journal = {Building and Environment},
  volume  = {221},
  pages   = {109345},
  year    = {2022},
  doi     = {10.1016/j.buildenv.2022.109345}
}

@article{kiya2022image,
  author  = {Kiya, Hitoshi and Iijima, Ryota and April-Pyone, Maung Maung and Kinoshita, Yuma},
  title   = {Image and Model Transformation with Secret Key for {Vision Transformer}},
  journal = {{IEICE} Transactions on Information and Systems},
  volume  = {E106-D},
  number  = {1},
  pages   = {2--11},
  year    = {2023},
  doi     = {10.1587/transinf.2022MUI0001}
}

@article{sirichotedumrong2019privacy,
  author  = {Sirichotedumrong, Warit and Kinoshita, Yuma and Kiya, Hitoshi},
  title   = {Pixel-Based Image Encryption without Key Management for Privacy-Preserving {Deep Neural Networks}},
  journal = {{IEEE} Access},
  volume  = {7},
  pages   = {177844--177855},
  year    = {2019},
  doi     = {10.1109/ACCESS.2019.2959017}
}

@inproceedings{liu2016deepfashion,
  author    = {Liu, Ziwei and Luo, Ping and Qiu, Shi and Wang, Xiaogang and Tang, Xiaoou},
  title     = {{DeepFashion}: Powering Robust Clothes Recognition and Retrieval with Rich Annotations},
  booktitle = {2016 {IEEE} Conference on Computer Vision and Pattern Recognition (CVPR)},
  pages     = {1096--1104},
  year      = {2016},
  doi       = {10.1109/CVPR.2016.124}
}

@inproceedings{deng2009imagenet,
  author = {Deng, Jia and Dong, Wei and Socher, Richard and Li, Li-Jia and Li, Kai and Li, Fei-Fei},
  title     = {{ImageNet}: A Large-Scale Hierarchical Image Database},
  booktitle = {2009 {IEEE} Conference on Computer Vision and Pattern Recognition (CVPR)},
  pages     = {248--255},
  year      = {2009},
  doi       = {10.1109/CVPR.2009.5206848}
}

\end{document}